\documentclass[10pt,twocolumn,letterpaper]{article}

\usepackage[pagenumbers]{cvpr}
\usepackage{stfloats}
\usepackage[accsupp]{axessibility}  
\definecolor{cvprblue}{rgb}{0.21,0.49,0.74}
\usepackage[pagebackref,breaklinks,colorlinks,allcolors=cvprblue]{hyperref}

\def\paperID{*****} 
\def\confName{CVPR}
\def\confYear{2026}

\title{Why CNN Features Are not Gaussian: A Statistical Anatomy of Deep Representations}

\author{
 David Chapman \quad Parniyan Farvardin \\
University of Miami\\
{\tt\small dchapman@cs.miami.edu, pxf291@miami.edu}
}
\begin{document}

\maketitle

\begin{abstract}
Deep convolutional neural networks (CNNs) are commonly analyzed through geometric and linear–algebraic perspectives, yet the statistical distribution of their internal feature activations remains poorly understood. In many applications, deep features are implicitly treated as Gaussian when modeling densities. In this work, we empirically examine this assumption and show that it does not accurately describe the distribution of CNN feature activations. Through a systematic study across multiple architectures and datasets, we find that the feature activations deviate substantially from Gaussian and are better characterized by Weibull and related long-tailed distributions. We further introduce a novel Discretized Characteristic Function Copula (DCF-Copula) method to model multivariate feature dependencies. We find that tail-length increases with network depth and that upper-tail dependence emerges between feature pairs.  These statistical findings are not consistent with the Central Limit Theorem, and are instead indicative of a Matthew process that progressively concentrates semantic signal within the tails. These statistical findings suggest that CNNs are excellent at noise reduction, yet poor at outlier removal tasks.  We recommend the use of long-tailed upper-tail-dependent priors as opposed to Gaussian priors for accurately CNN deep feature density.  Code available at \href{https://github.com/dchapman-prof/DCF-Copula}{https://github.com/dchapman-prof/DCF-Copula}

\end{abstract}

\vspace{-2pt}

\section{Introduction}
\label{sec:intro}

Understanding the statistical distribution of CNN deep feature activations remains an important yet largely unexplored problem.
Mechanistic interpretability aims to reverse engineer deep learning models in order to understand how their internal representations are formed \cite{elhage2022toy,elhage2021mathematical,wang2022interpretability}. While significant progress has been made in understanding the \textit{semantic meaning} of deep vision features, the \textit{statistical structure} of these activations has received far less attention \cite{allen2023towards, elhage2022toy, noh2015learning, zhang2018interpretable, zhang2021understanding, feldman2020discovering, feldman2020short, vladimirova2019bayesian}. 
Our goal is to reverse engineer these feature distributions, including marginal and interdependence behavior using exploratory analysis and confirmatory tests.

Understanding this distribution matters because many methods implicitly assume Gaussian priors \cite{kingma2013auto, kingma2016improved, majurski2024method, zhu2022boosting, lee2018simple, rippel2021modeling}.  If these assumptions do not hold, learning methods based upon them may be unpredictable or sub-optimal. 

If Gaussian priors were realistic for activations, then Anomaly Detection (AD) would be inherently easy; it is not \cite{nalisnick2018deep, lee2018simple, liu2020energy, zhu2022boosting, le2021perfect, jiang2023detecting, zhang2021understanding}.
Gaussian priors motivate simple density and outlier-removal strategies (e.g., ellipsoidal level sets).  Just detect the outliers, call them anomalies.  Clearly things are not so simple, and successful AD methods have often avoided this approach \cite{hojjati2024self, zhou2023anomalyclip, jeong2023winclip, wan2026dcs}, but why?

Our empirical analysis provides new clues about this discrepancy. In particular, long-tailed distributions with upper-tail dependence are poorly matched to classical \textit{outlier detection} assumptions.  Under Gaussian assumptions, outlier removal should improve performance.  We find the opposite, outlier removal catastrophically destroys performance.


Our explanation is that deep learning models attempt to learn the long-tailed distribution of the natural underlying image  statistics.  The distribution of object parts in imagery naturally follows a power law \cite{feldman2020discovering, feldman2020short}.  We find that CNN must \textit{learn} this distribution, starting with an uninformed prior, and layer-by-layer increasing the tail length. 

We propose the use of a Weibull marginal to represent deep feature marginals with a range of tail lengths.  We analyze feature interdependence using a novel DCF-Copula approach that is highly expressive, and we observe upper tail dependence.  We argue that most of the semantic signal in deep features is represented by the tail, and thus we recommend the consideration of long-tailed upper-tail dependent priors.
Our \underline{contributions} are as follows.

\begin{itemize}
\item Systematic exploratory and confirmatory analysis of the CNN deep feature marginals and interdependence.
\item Novel DCF-Copula analysis framework that is more expressive than Archimedean copulas.
\item Observe long-tailed, and upper-tail-dependent behavior with tail length increasing with depth.
\item Outlier removal destroys performance, but noise suppression improves performance due to feature distribution.
\end{itemize}

\section{DCF-Copula Methodology}
\label{sec:methodology}



We now define the DCF-Copula methodology that is used to empirically estimate the statistical behavior of deep CNN features.
Copula analysis separates the marginal distributions of random variables from their dependence structure.  
Given random variables $(X_1,\dots,X_D)$ with cumulative distributions $(F_1,\dots,F_D)$. 
The copula $C$ is defined as the joint cumulative distribution of these transformed variables

\begin{equation}
C(y_1,\dots,y_D) = Pr[Y_1 \le y_1,\dots,Y_D \le y_D],
\end{equation}









Typical Archimedean copulas are inflexible, and represent density using parametric methods.  Our DCF-Copula method is highly expressive because we apply the Method of Orthogonal Moments (MOM) to represent copula density, thereby capturing all multivariate non-linear statistical dependencies.
Given basis functions $\phi_t(\cdot)$, population and sample moments are

\begin{equation}
\mu_t = E[\phi_t(x)], \qquad
\hat{\mu}_t = \frac{1}{N}\sum_{i=1}^N \phi_t(x_i).
\end{equation}

For multivariate variables $Y_1,\dots,Y_D$, joint moments capture cross-feature dependence

\begin{equation}
\mu_T = E\!\left(\prod_{d=1}^{D}\phi_{T_d}(Y_d)\right),
\quad
\hat{\mu}_T =
\frac{1}{N}\sum_{i=1}^N
\prod_{d=1}^{D}\phi_{T_d}(Y_{d,i}).
\end{equation}



DCF-Copula reconstructs density using a discretized version of the Generalized Characteristic Function 
which requires far fewer terms than the traditional characteristic function techniques. 
If $\phi_t$ is an orthogonal basis, the moments correspond to transform coefficients

\begin{equation}
\mu_t = E[\phi_t(y)] = \int c(y)\phi_t(y)dy.
\end{equation}

Using a finite expansion, the copula density can be obtained as

\begin{equation}
c(y) = \sum_{t=1}^K \mu_t\phi_t(y), \qquad
\hat{c}(y) = \sum_{t=1}^K \hat{\mu}_t\phi_t(y).
\end{equation}




We find the normalized Legendre polynomials and real-valued Fourier harmonics to be a suitable basis function series.  We discuss these functions and their desirable properties in greater detail in Appendix ~\ref{app:orthogonal}.
The multivariate basis series is defined as the product of the univariate basis function $\phi_{T_d}$ as follows.


\begin{equation}
\Phi_T(\vec{y}) = \prod_{d=1}^D \phi_{T_d}(y_d).
\end{equation}

The empirical copula density is then estimated as

\begin{equation}
\hat{c}(\vec{y}) =
\sum_{T\in \mathbb{Z}_K^D}
\hat{\mu}_T \Phi_T(\vec{y}).
\end{equation}


\section{Tail-Length Parameterization}

We perform tail analysis in order to quantify the long-tailed nature of the CNN deep features.  Toward this aim we use the Weibull distribution, as this allows for a range of tail lengths from sub-Gaussian to long-tailed.
The PDF and CDF of the Weibull distribution are defined as the following.

\begin{equation}
f(x;\theta,k) = \begin{cases} \displaystyle
      \theta k \left( \theta x \right)^{k-1} e^{-\left( \theta x \right)^k} & x \ge 0 \\
      0 &  x < 0  
   \end{cases}
\end{equation}

\begin{equation}
F(x;\theta,k) = \begin{cases} \displaystyle
      1 - e^{-\left( \theta x \right)^k} & x \ge 0 \\
      0 &  x < 0  
   \end{cases}
\end{equation}

Where $\theta$ is the tail parameter, and $k$ is the shape parameter.  In particular the tail parameter $\theta$ is indicative of the tail-length with $\theta \le 0.5$ indicating sub-Gaussian, $\theta \le 1$ indicating sub-exponential, and $\theta > 1$ indicating long-tailed.


For tail analysis, we only consider samples that exceed the $99^{th}$ percentile which is a common choice for extreme value theory.  
We define $\ddot{x} \subseteq X$ as the high value subset  as follows, 

\begin{equation}
\begin{aligned}
\ddot{x} \; = \; \{ \; \ddot{x}_i \; : \; \ddot{x}_i \in X \text{ and } \ddot{x}_i > u\; \} \\[5pt]
\text{where} \quad  Pr[X \le u] = 0.99
\end{aligned}
\end{equation}

Our loss function for fitting the tail is the 1-Wasserstein distance, because it is a well-behaved distance metric for comparing partial distributions.  The 1-Wasserstein distance can be calculated by comparing the CDF of the observed samples $\ddot{y}$ with that of the theoretical Weibull distribution $F(\ddot{x};\theta;k)$ in order to determine the optimal Weibull tail parameter of the deep feature marginals.


\begin{equation}
\underset{\theta, \ \! k}{argmin} \; \; \sum_{i=1}^n \; \big|\big| \; \ddot{y}_i - F( \ \! \ddot{x}_i \ \! ;\theta,k) \;\big|\big|_1 
\end{equation}

\begin{figure*}[t]
\centering
\includegraphics[width=\textwidth]{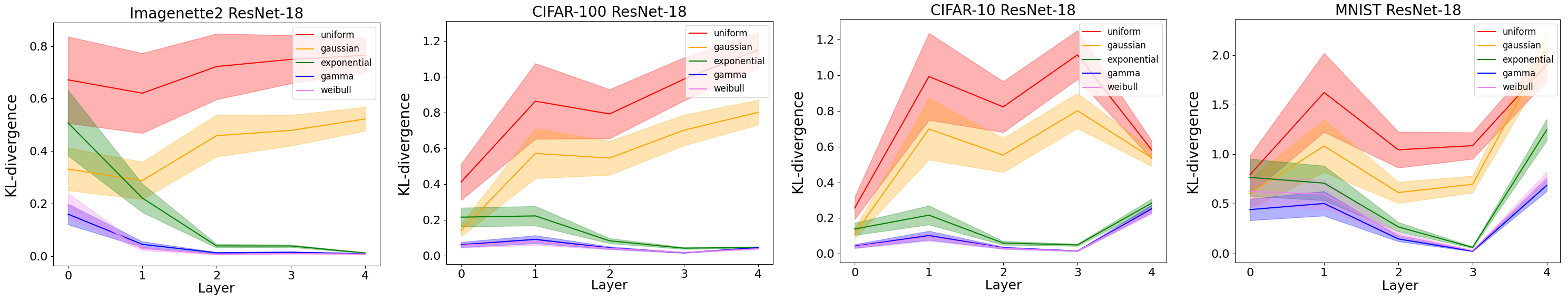}
\caption{Overall goodness of fit of parametric marginal distributions, with shaded region showing $2 \sigma$ error bars.}
\label{figures:fit}
\vspace{5pt}
\includegraphics[width=\textwidth]{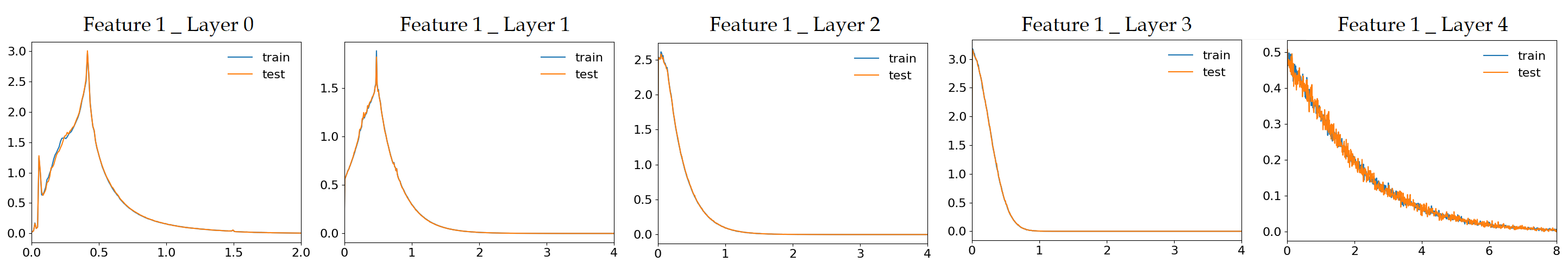}
\caption{Example histograms of ResNet-18 feature density for extracted layers on Imagenette2.}
\label{figures:marginal}

\vspace{-5pt}
\end{figure*}

\section{Results}
\label{sec:results}

\begin{figure}[b]
\centering
\includegraphics[width=0.375\textwidth]{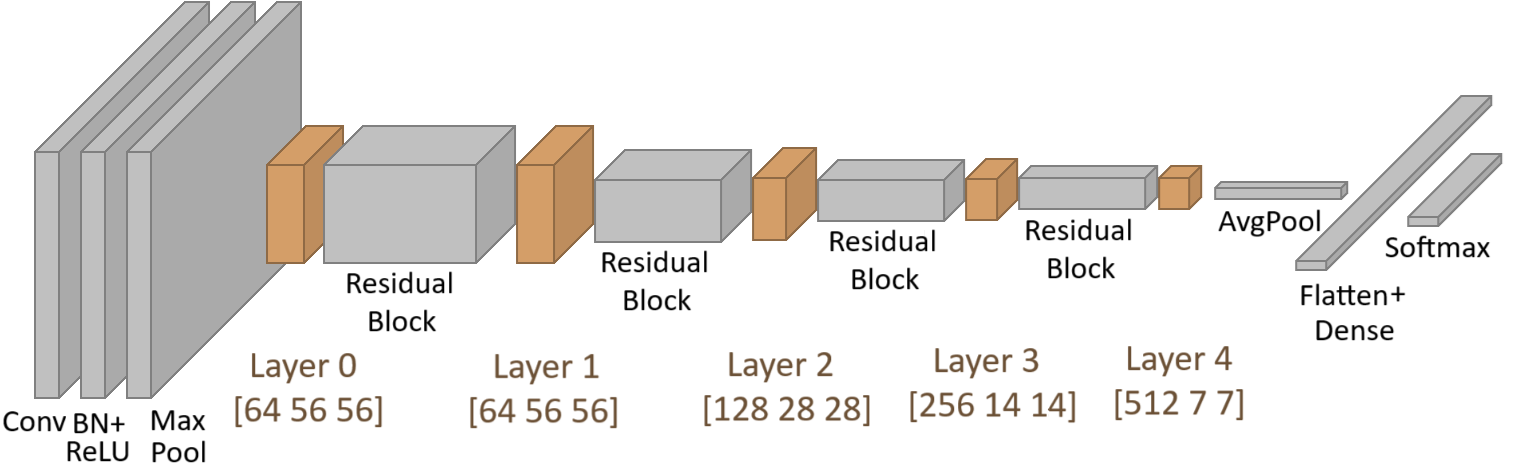}
\caption{ResNet-18, orange shows extracted features.}
\label{figures:cnn}
\end{figure}

We analyze the deep features for three CNNs, ResNet-18 \cite{he2016deep}, ResNet-50 \cite{he2016deep}, and VGG-19 \cite{simonyan2014very} across four image classification datasets: MNIST \cite{deng2012mnist}, CIFAR-10 \citep{krizhevsky2009learning}, CIFAR-100 \citep{krizhevsky2009learning}, and Imagenette2 \cite{deng2009imagenet, Howard_Imagenette_2019}.  Figure \ref{figures:cnn} shows the extracted features ResNet-18 spanning multiple layer depths; similar features were extracted for other models.

Figure \ref{figures:marginal} shows examples of univariate histograms of ResNet-18 on Imagenette2. Plots for other datasets are shown in Appendix ~\ref{app:marginal}. All plots show a prominent tail-shape.  The early layers show some remnants of a bell-curve due to the lingering distribution of input colors.  But by the third layer onward, there is no more bell-curve, and the entire histogram appears tail-like.  By the final layer, the tail is very long and prominent.  We see similar behavior on the other datasets ~\ref{app:marginal}.
These visual results already suggest that the deep semantic features form a prominent tail-like distribution that becomes increasing prominent with depth.  
Instead of the Central Limit Theorem (increasing bell-shape), this is a Matthew effect (increasing tail-shape) indicating a winners-compounding process (i.e. semantic signals become more semantic with depth).

The second step is confirmatory analysis to obtain a suitable distribution.  In Figure \ref{figures:fit} we analyze five standard distributions: Uniform, Gaussian, Exponential, Gamma, and Weibull.  Similar tests are performed for ResNet-50 and VGG-19 in Appendix ~\ref{app:marginal}.  
Numeric details of this test are given in Appendix ~\ref{app:fit}.  This hypothesis test confirms with high significance that for all datasets and models, the Uniform and Gaussian distributions fit very poorly after the first layer, whereas the Exponential, Gamma and Weibull distributions are a much better fit.  Moreover, it is highly significant that the Gamma and Weibull distributions outperform the Exponential distribution.
From here onward we assume a Weibull distribution, because it is not only a good distributional fit, but it can also model a range of tail behaviors by adjusting its $\theta$ parameter.

\begin{figure*}[t]
\centering
\includegraphics[width=\linewidth]{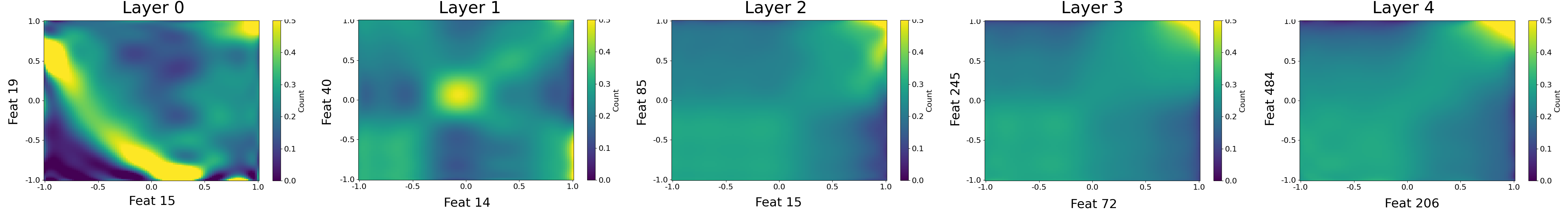}
\caption{DCF-Copula density between random feature pairs for ResNet-18 on Imagenette2.}
\label{figures:copula}
\vspace{10pt}
\includegraphics[width=\linewidth]{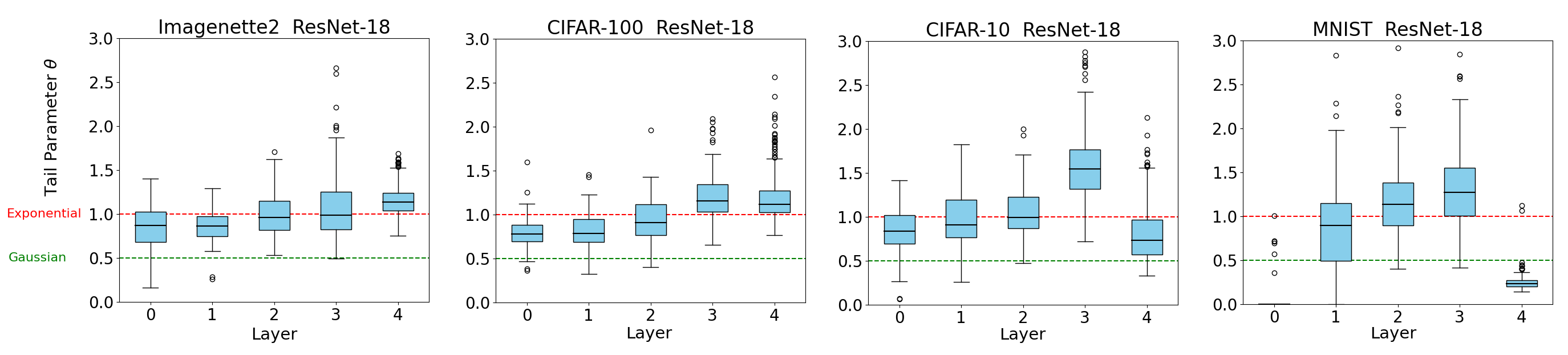}
\caption{Estimated Weibull tail parameter $\theta$ for ResNet-18 features across layers. Larger values correspond to heavier-tailed activation distributions. Deeper layers exhibit stronger tail behavior, particularly for more complex datasets.}
\label{figures:tail}
\vspace{-5pt}
\end{figure*}

Our next goal is to determine just how \textit{long-tailed} the feature distribution really is.  Figure \ref{figures:tail} shows the estimated Weibull $\theta$ parameter across all layers for the upper $99^{th}$ percentile of observations for ResNet-18.  Additional plots for ResNet-50 and VGG-19 are shown in Appendix ~\ref{app:marginal}.  We see a Matthew effect, where the length of the $\theta$ parameter increases with each successive layer.  Although for the \textit{easy} datasets (MNIST and CIFAR-10), $\theta$ sharply drops off in the last layer.  But $\theta$ does not drop in the \textit{hard} datasets (Imagenette2 and CIFAR-100).

\label{sec:tail}


Also notable, is that although the tail length increases with depth until long-tailed $\theta>1$ it never reaches a power-law $\theta \rightarrow \infty$.  Natural image statistics follow power-laws, so why do deep CNNs exhibit a more modest long-tail?
Our explanation is that the neural network never fully learns the full spectrum of semantic concepts within the image, only enough concepts for the subtasks of classification, and only after many layers.  After each layer the network represents more and more semantic signal leading to a longer and longer tail (Matthew effect).  But it never learns the full power-law spectrum within the natural image.  Moreover, if the classification task is \textit{easy} (CIFAR-10, MNIST), it abruptly shortens the tail just before classification.

Our next step is to perform non-parametric Copula analysis of the feature inter-dependence.  Table \ref{tab:copula_compare} shows that our proposed DCF-Copula method greatly outperforms a range of comparable methods to model statistical interdependence including the Archimedean copulas (AMH, Clayton, Frank, Gumbel, Joe) as well as characteristic functions (ECF).  Additional details of the experiment are given in Appendix ~\ref{app:confidence} additional plots in Appendix ~\ref{app:marginal}.

\begin{table}[b]
\centering
\caption{Copula density loss for ResNet-18 on Imagenette2.}
\renewcommand{\arraystretch}{1.1}
\setlength{\tabcolsep}{4pt}
\begin{tabular}{lccccc}
\hline
 & L0 & L1 & L2 & L3 & L4 \\
\hline
DCF-Legendre & 1.1883 & 1.3757 & \textbf{1.3794} & \textbf{1.3823} & \textbf{1.3711} \\
DCF-Fourier  & \textbf{1.1878} & \textbf{1.3748} & 1.3796 & 1.3825 & 1.3716 \\
\hline
ECF \cite{nolan2013elliptically}     & 2.1029 & 2.0450 & 2.1529 & 2.0635 & 1.4537 \\
AMH \cite{ali1978class}     & 1.5329 & 1.4933 & 1.4953 & 1.4969 & 1.5002 \\
Clayton \cite{clayton1978model} & 1.4469 & 1.4756 & 1.4755 & 1.4806 & 1.4631 \\
Frank \cite{frank1979simultaneous} &  1.4779 & 1.4818 & 1.4825 & 1.4869 & 1.4762
 \\
Gumbel \cite{gumbel1960bivariate} & 1.4749 & 1.4836 &  1.4845 & 1.4888 & 1.4808
 \\
Joe \cite{joe1994multivariate} & 1.5020 & 1.4867 & 1.4883 & 1.4917 & 1.4890
 \\
\hline
\end{tabular}
\label{tab:copula_compare}
\end{table}

Figure \ref{figures:copula} shows example bivariate copula plots for random feature pairs.  We see that the first two layers show remnants of the input pixel distribution, but all subsequent layers show a bright spot at (1,1) (upper right).  This indicates an upper tail dependence, that features are uncorrelated within their normal ranges, but highly correlated in the tail regime.  We believe this is due to \textit{polysemanticity} \cite{elhage2022toy}, the observation that a single neuron feature will fire due to multiple semantic concepts.  Similarly, a single highly semantic region of an image (i.e. object part) may cause multiple neurons to strongly fire simultaneously, thereby creating an upper tail dependence.

\begin{figure}[htb]
\centering
\includegraphics[width=0.5\textwidth]{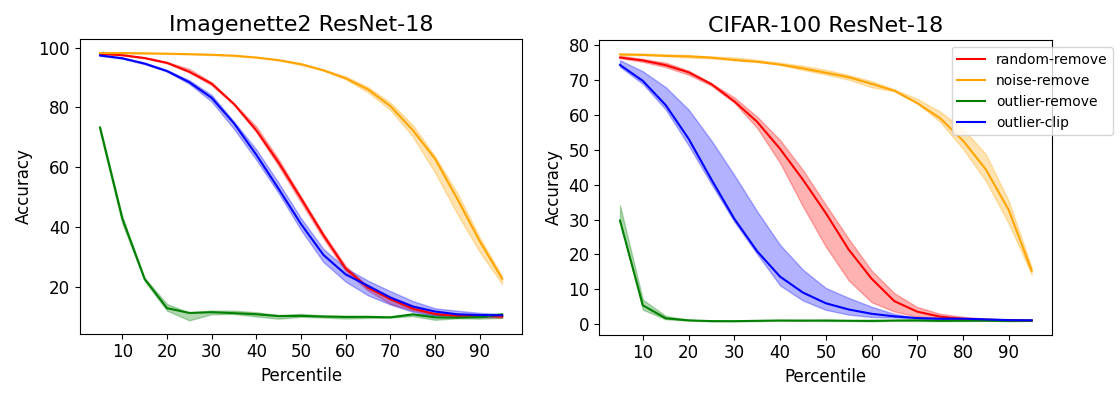}
\caption{Accuracy vs removal of noise and outlier signals.}
\label{figures:noise}
\vspace{-8pt}
\end{figure}

We conclude our analysis with Figure \ref{figures:noise}, showing what happens if one unintentionally confuses the tail of the distribution with outliers. Removing the high-percentile features (green),  catastrophically destroys the accuracy.  But the model can tolerate removal of low percentile features (yellow).  Even clipping the high percentile features (blue) underperforms zeroing an equal number of random features (red).  Clearly the tail exhibits an important signal, whereas the head of the distribution is irrelevant for classification.

\section{Conclusion}
\label{sec:conclusion}

Our results bring a new understanding to the statistical distribution of deep CNN features.  We clearly show that the tail of the distribution contains the most important signal.  We observe that deep CNN features follow a Matthew process, as the tail length increases layer-by-layer.  We believe this is due to the models increasing ability to represent natural image statistics with increasing depth.  Upper-tail dependence may be due to polysemanticity, or natural image statistics, or both.  Many advanced learning methods incorporate statistical feature priors, with the Gaussian prior being a common choice.
Matthew process behavior is not consistent with the Central Limit Theorem.
This is why the CNN features are not Gaussian, and we recommend the careful consideration of long-tailed distributions instead.




\section*{Acknowledgement}

This work is supported by the Frost Institute for Data Science and Computing.

{
    \small
    \bibliographystyle{ieeenat_fullname}
    \bibliography{main}

@String(ICPR = {Int. Conf. Pattern Recog.})

@String(ICPR  = {ICPR})

@inproceedings{allen2023towards,
  title={Towards Understanding Ensemble, Knowledge Distillation and Self-Distillation in Deep Learning},
  author={Allen-Zhu, Zeyuan and Li, Yuanzhi},
  booktitle={The Eleventh International Conference on Learning Representations},
  year={2023}
}

@inproceedings{jiang2023detecting,
  title={Detecting out-of-distribution data through in-distribution class prior},
  author={Jiang, Xue and Liu, Feng and Fang, Zhen and Chen, Hong and Liu, Tongliang and Zheng, Feng and Han, Bo},
  booktitle={International Conference on Machine Learning},
  pages={15067--15088},
  year={2023},
  organization={PMLR}
}

@article{le2021perfect,
  title={Perfect density models cannot guarantee anomaly detection},
  author={Le Lan, Charline and Dinh, Laurent},
  journal={Entropy},
  volume={23},
  number={12},
  pages={1690},
  year={2021},
  publisher={MDPI}
}

@article{lee2018simple,
  title={A simple unified framework for detecting out-of-distribution samples and adversarial attacks},
  author={Lee, Kimin and Lee, Kibok and Lee, Honglak and Shin, Jinwoo},
  journal={Advances in neural information processing systems},
  volume={31},
  year={2018}
}

@inproceedings{liu2020energy,
 author = {Liu, Weitang and Wang, Xiaoyun and Owens, John and Li, Yixuan},
 booktitle = {Advances in Neural Information Processing Systems},
 editor = {H. Larochelle and M. Ranzato and R. Hadsell and M.F. Balcan and H. Lin},
 pages = {21464--21475},
 publisher = {Curran Associates, Inc.},
 title = {Energy-based Out-of-distribution Detection},
 url = {https://proceedings.neurips.cc/paper_files/paper/2020/file/f5496252609c43eb8a3d147ab9b9c006-Paper.pdf},
 volume = {33},
 year = {2020}
}

@inproceedings{rippel2021modeling,
  title={Modeling the distribution of normal data in pre-trained deep features for anomaly detection},
  author={Rippel, Oliver and Mertens, Patrick and Merhof, Dorit},
  booktitle={2020 25th International Conference on Pattern Recognition (ICPR)},
  pages={6726--6733},
  year={2021},
  organization={IEEE}
}

@inproceedings{zhang2021understanding,
  title={Understanding failures in out-of-distribution detection with deep generative models},
  author={Zhang, Lily and Goldstein, Mark and Ranganath, Rajesh},
  booktitle={International Conference on Machine Learning},
  pages={12427--12436},
  year={2021},
  organization={PMLR}
}

@inproceedings{zhu2022boosting,
 author = {Zhu, Yao and Chen, YueFeng and Xie, Chuanlong and Li, Xiaodan and Zhang, Rong and Xue\textquotesingle , Hui and Tian, Xiang and zheng, bolun and Chen, Yaowu},
 booktitle = {Advances in Neural Information Processing Systems},
 editor = {S. Koyejo and S. Mohamed and A. Agarwal and D. Belgrave and K. Cho and A. Oh},
 pages = {20758--20769},
 publisher = {Curran Associates, Inc.},
 title = {Boosting Out-of-distribution Detection with Typical Features},
 url = {https://proceedings.neurips.cc/paper_files/paper/2022/file/82b0c1b954b6ef9f3cfb664a82b201bb-Paper-Conference.pdf},
 volume = {35},
 year = {2022}
}

@article{kingma2013auto,
  title={Auto-encoding variational bayes},
  author={Kingma, Diederik P},
  journal={arXiv preprint arXiv:1312.6114},
  year={2013}
}

@article{kingma2016improved,
  title={Improved variational inference with inverse autoregressive flow},
  author={Kingma, Durk P and Salimans, Tim and Jozefowicz, Rafal and Chen, Xi and Sutskever, Ilya and Welling, Max},
  journal={Advances in neural information processing systems},
  volume={29},
  year={2016}
}

@inproceedings{majurski2024method,
  title={A Method of Moments Embedding Constraint and its Application to Semi-Supervised Learning},
  author={Majurski, Michael and Menon, Sumeet and Favardin, Parniyan and Chapman, David},
  booktitle={Proceedings of the IEEE/CVF Conference on Computer Vision and Pattern Recognition},
  pages={7809--7818},
  year={2024}
}

@article{deng2012mnist,
  title={The mnist database of handwritten digit images for machine learning research},
  author={Deng, Li},
  journal={IEEE Signal Processing Magazine},
  volume={29},
  number={6},
  pages={141--142},
  year={2012},
  publisher={IEEE}
}

@article{krizhevsky2009learning,
  title={Learning multiple layers of features from tiny images},
  author={Krizhevsky, Alex and Hinton, Geoffrey and others},
  year={2009},
  publisher={Toronto, ON, Canada}
}

@software{Howard_Imagenette_2019,
    title={Imagenette: A smaller subset of 10 easily classified classes from Imagenet},
    author={Jeremy Howard},
    year={2019},
    month={March},
    publisher = {GitHub},
    url = {https://github.com/fastai/imagenette}
}

@INPROCEEDINGS{deng2009imagenet,
  author={Deng, Jia and Dong, Wei and Socher, Richard and Li, Li-Jia and Kai Li and Li Fei-Fei},
  booktitle={2009 IEEE Conference on Computer Vision and Pattern Recognition}, 
  title={ImageNet: A large-scale hierarchical image database}, 
  year={2009},
  volume={},
  number={},
  pages={248-255},
  doi={10.1109/CVPR.2009.5206848}}

@inproceedings{he2016deep,
  title={Deep residual learning for image recognition},
  author={He, Kaiming and Zhang, Xiangyu and Ren, Shaoqing and Sun, Jian},
  booktitle={Proceedings of the IEEE conference on computer vision and pattern recognition},
  pages={770--778},
  year={2016}
}

@article{simonyan2014very,
  title={Very deep convolutional networks for large-scale image recognition},
  author={Simonyan, Karen and Zisserman, Andrew},
  journal={arXiv preprint arXiv:1409.1556},
  year={2014}
}

@inproceedings{vladimirova2019bayesian,
  title={Understanding priors in Bayesian neural networks at the unit level},
  author={Vladimirova, Mariia and Verbeek, Jakob and Mesejo, Pablo and Arbel, Julyan},
  booktitle={International Conference on Machine Learning},
  pages={6458--6467},
  year={2019},
  organization={PMLR}
}

@inproceedings{zhang2018interpretable,
  title={Interpretable convolutional neural networks},
  author={Zhang, Quanshi and Wu, Ying Nian and Zhu, Song-Chun},
  booktitle={Proceedings of the IEEE conference on computer vision and pattern recognition},
  pages={8827--8836},
  year={2018}
}

@inproceedings{feldman2020short,
  title={Does learning require memorization? a short tale about a long tail},
  author={Feldman, Vitaly},
  booktitle={Proceedings of the 52nd Annual ACM SIGACT Symposium on Theory of Computing},
  pages={954--959},
  year={2020}
}

@article{feldman2020discovering,
  title={What neural networks memorize and why: Discovering the long tail via influence estimation},
  author={Feldman, Vitaly and Zhang, Chiyuan},
  journal={Advances in Neural Information Processing Systems},
  volume={33},
  pages={2881--2891},
  year={2020}
}

@article{nalisnick2018deep,
  title={Do deep generative models know what they don't know?},
  author={Nalisnick, Eric and Matsukawa, Akihiro and Teh, Yee Whye and Gorur, Dilan and Lakshminarayanan, Balaji},
  journal={arXiv preprint arXiv:1810.09136},
  year={2018}
}

@article{nolan2013elliptically,
  title={Multivariate elliptically contoured stable distributions: theory and estimation},
  author={Nolan, John},
  journal={Computational Statistics},
  volume={28},
  number={5},
  pages={2067--2089},
  year={2013}
}

@inproceedings{noh2015learning,
  title={Learning deconvolution network for semantic segmentation},
  author={Noh, Hyeonwoo and Hong, Seunghoon and Han, Bohyung},
  booktitle={Proceedings of the IEEE international conference on computer vision},
  pages={1520--1528},
  year={2015}
}

@article{ali1978class,
  title={A class of bivariate distributions including the bivariate logistic},
  author={Ali, Mir M and Mikhail, NN and Haq, M Safiul},
  journal={Journal of multivariate analysis},
  volume={8},
  number={3},
  pages={405--412},
  year={1978},
  publisher={Elsevier}
}

@article{clayton1978model,
  title={A model for association in bivariate life tables and its application in epidemiological studies of familial tendency in chronic disease incidence},
  author={Clayton, David G},
  journal={Biometrika},
  volume={65},
  number={1},
  pages={141--151},
  year={1978},
  publisher={Oxford University Press}
}

@article{joe1994multivariate,
  title={Multivariate extreme-value distributions with applications to environmental data},
  author={Joe, Harry},
  journal={Canadian Journal of Statistics},
  volume={22},
  number={1},
  pages={47--64},
  year={1994},
  publisher={Wiley Online Library}
}

@article{gumbel1960bivariate,
  title={Bivariate exponential distributions},
  author={Gumbel, Emil J},
  journal={Journal of the American Statistical Association},
  volume={55},
  number={292},
  pages={698--707},
  year={1960},
  publisher={Taylor \& Francis}
}

@article{frank1979simultaneous,
  title={On the simultaneous associativity of F (x, y) and x+ y- F (x, y)},
  author={Frank, Maurice J},
  journal={Aequationes mathematicae},
  volume={19},
  pages={194--226},
  year={1979},
  publisher={Springer}
}

@article{elhage2022toy,
  title={Toy models of superposition},
  author={Elhage, Nelson and Hume, Tristan and Olsson, Catherine and Schiefer, Nicholas and Henighan, Tom and Kravec, Shauna and Hatfield-Dodds, Zac and Lasenby, Robert and Drain, Dawn and Chen, Carol and others},
  journal={arXiv preprint arXiv:2209.10652},
  year={2022}
}

@article{elhage2021mathematical,
  title={A mathematical framework for transformer circuits},
  author={Elhage, Nelson and Nanda, Neel and Olsson, Catherine and Henighan, Tom and Joseph, Nicholas and Mann, Ben and Askell, Amanda and Bai, Yuntao and Chen, Anna and Conerly, Tom and others},
  journal={Transformer Circuits Thread},
  volume={1},
  number={1},
  pages={12},
  year={2021}
}

@article{wang2022interpretability,
  title={Interpretability in the wild: a circuit for indirect object identification in gpt-2 small},
  author={Wang, Kevin and Variengien, Alexandre and Conmy, Arthur and Shlegeris, Buck and Steinhardt, Jacob},
  journal={arXiv preprint arXiv:2211.00593},
  year={2022}
}

@article{hojjati2024self,
  title={Self-supervised anomaly detection in computer vision and beyond: A survey and outlook},
  author={Hojjati, Hadi and Ho, Thi Kieu Khanh and Armanfard, Narges},
  journal={Neural Networks},
  volume={172},
  pages={106106},
  year={2024},
  publisher={Elsevier}
}

@article{zhou2023anomalyclip,
  title={Anomalyclip: Object-agnostic prompt learning for zero-shot anomaly detection},
  author={Zhou, Qihang and Pang, Guansong and Tian, Yu and He, Shibo and Chen, Jiming},
  journal={arXiv preprint arXiv:2310.18961},
  year={2023}
}

@inproceedings{jeong2023winclip,
  title={Winclip: Zero-/few-shot anomaly classification and segmentation},
  author={Jeong, Jongheon and Zou, Yang and Kim, Taewan and Zhang, Dongqing and Ravichandran, Avinash and Dabeer, Onkar},
  booktitle={Proceedings of the IEEE/CVF conference on computer vision and pattern recognition},
  pages={19606--19616},
  year={2023}
}

@article{wan2026dcs,
  title={DCS: A Zero-Shot Anomaly Detection Framework with DINO-CLIP-SAM Integration},
  author={Wan, Yan and Lang, Yingqi and Yao, Li},
  journal={Applied Sciences},
  volume={16},
  number={4},
  pages={1836},
  year={2026},
  publisher={MDPI}
}
}

\clearpage
\appendix

\appendix
\section{Appendix: Orthogonal Basis Functions}
\label{app:orthogonal}

\begin{figure}[hbt]
\centering
\includegraphics[width=0.5\textwidth]{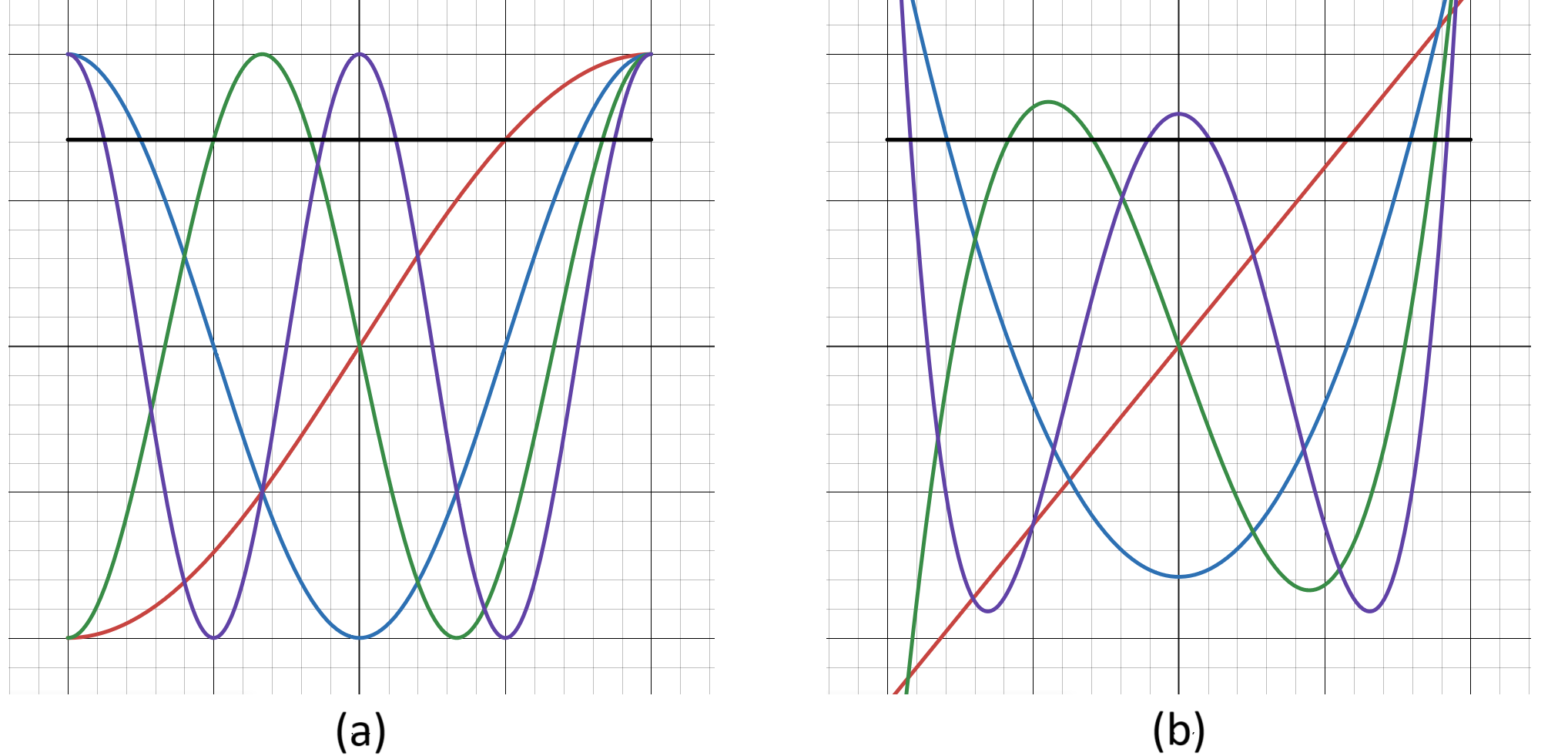}
\caption{a. Real-valued Fourier Series  b. Normalized Legendre Polynomials}
\label{figures:basis}
\end{figure}

DCF-Copula requires orthogonal basis functions with the following properties.

\begin{itemize}
\item Orthogonal over unit interval $(-1, 1)$
\item Real valued, exhibiting even and odd harmonics
\item Unit length $L_2$ norm over interval $(-1, 1)$
\item All non-constant moments exhibit zero integral over $(-1, 1)$
\end{itemize}

We propose a specific normalized form of the Legendre polynomials, as well as a real-valued form of the Fourier series as seen in figure \ref{figures:basis}.

\subsection{Normalized Legendre Polynomials}

The Legendre Polynomials (figure \ref{figures:basis}b) are a set of real-valued orthogonal basis functions over the target interval $(-1, 1)$ with several desirable properties.  Unlike the Chebyshev polynomials, the Legendre polynomials exhibit zero integral over the interval $(-1, 1)$, except trivially for the constant polynomial $P_0$. 
The following polynomials can be generated efficiently using Bonnet's recurrence. The Legendre polynomials in this form do not exhibit unit-length $L_2$ norm over the interval $(-1, 1)$, as such normalize these Legendre polynomials based on their $L_2$ as follows.

\begin{equation}
\phi_t(y) = \frac{P_t(y)}{||P_t||_2} \quad \text{where} \quad ||P_t||_2 = \sqrt{\int_{-1}^{1} P_t^2(y) \; dy}
\label{legmom}
\end{equation}



\subsubsection{Real-valued Fourier Series}

As our sample is real-valued, one can equivalently represent the Fourier series as a sum of real-valued $cos$ (even) and $sin$ (odd) harmonic terms.  Moreover, it is possible to simplify this to only $cos$ terms if one makes use of the trigonometric phase identity, which is given by the following. 



\begin{equation}
\begin{aligned}
\phi_0(y)&=\frac{\sqrt{2}}{2} \\
\phi_t(y)&=cos\left( t \frac{\pi}{2} (y-1) \right)
\end{aligned}
\label{fourmom}
\end{equation}

The real-valued Fourier basis functions in this form are shown in (figure \ref{figures:basis}a).  These basis functions also exhibit all of our required properties.

\section{Appendix: Details of Goodness of Fit}
\label{app:fit}

After fitting, the parametric models are evaluated using the KL-divergence between the parametric fit and the test histogram.
As such, the task presented is to determine how well each of the parametric models fit to the training histogram can approximate the empirical distribution of test samples as measured by a histogram.  
The shaded regions present $95\%$ confidence intervals of the layer-by-layer KL-divergence as estimated using Student's t-test.  A breakdown of the steps involved with this testing procedures are as follows.


\begin{enumerate}
    \item Compute the KL-divergence for the non-zeros samples of each filter $d$ within the target layer of $D$ filters, We denote this KL-divergence as $\mathrm{KL}_d$. This value measures how well the trained parametric model explains the test histogram of filter $d$.
    
    \item Compute the sample mean of KL-divergence values across all non-zero features in the layer.
   \begin{equation}
   \begin{aligned}
    \overline{\mathrm{KL}} = \frac{1}{D} \sum_{d=1}^{D} \mathrm{KL}_d
   \end{aligned}
   \end{equation}
    
    \item Estimate the standard error of the mean (SE), which quantifies the uncertainty in the estimated average KL-divergence where $s$ is the sample standard deviation of KL-divergences,
    \begin{equation}
    \begin{aligned}
    s = \frac{\sigma}{\sqrt{D}}, \quad \text{where } \sigma = \sqrt{ \frac{1}{D - 1} \sum_{d=1}^{D} (\mathrm{KL}_d - \overline{\mathrm{KL}})^2 }
    \end{aligned}
   \end{equation}
    
    \item Construct a $95\%$ confidence interval (CI) around the mean. Since $N \geq 64$, we approximate the $t$-distribution with the standard normal distribution. 
    \begin{equation}
    \begin{aligned}
    \mathrm{CI} = \overline{\mathrm{KL}} \pm z_{0.975} \cdot s, \quad \text{with } z_{0.975} \approx 1.96
    \end{aligned}
   \end{equation}
\end{enumerate}

\begin{figure*}[hbt]
\centering
\includegraphics[width=1.0\textwidth]{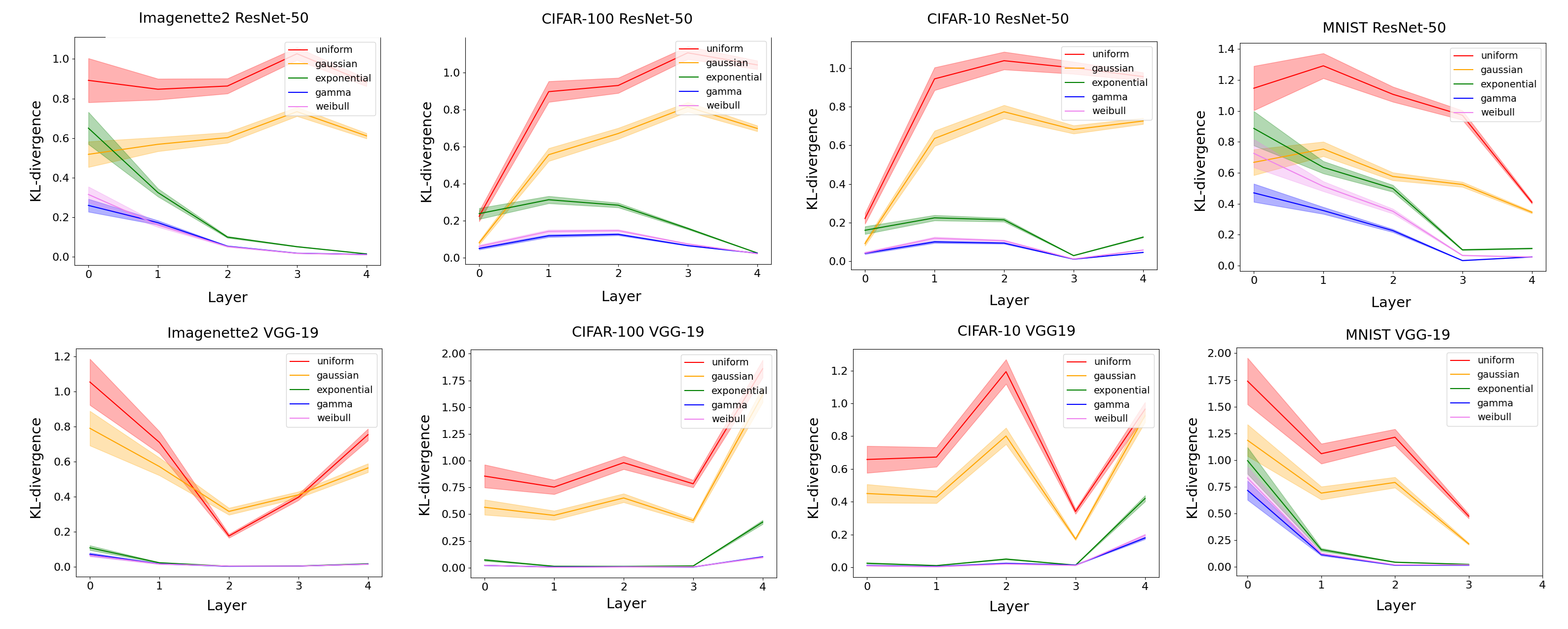}
\caption{Overall goodness of fit of parametric marginal distributions, with shaded regions showing $2\sigma$ error bars.}
\label{figures:fit_extra}
\vspace{20pt}

\includegraphics[width=1.0\textwidth]{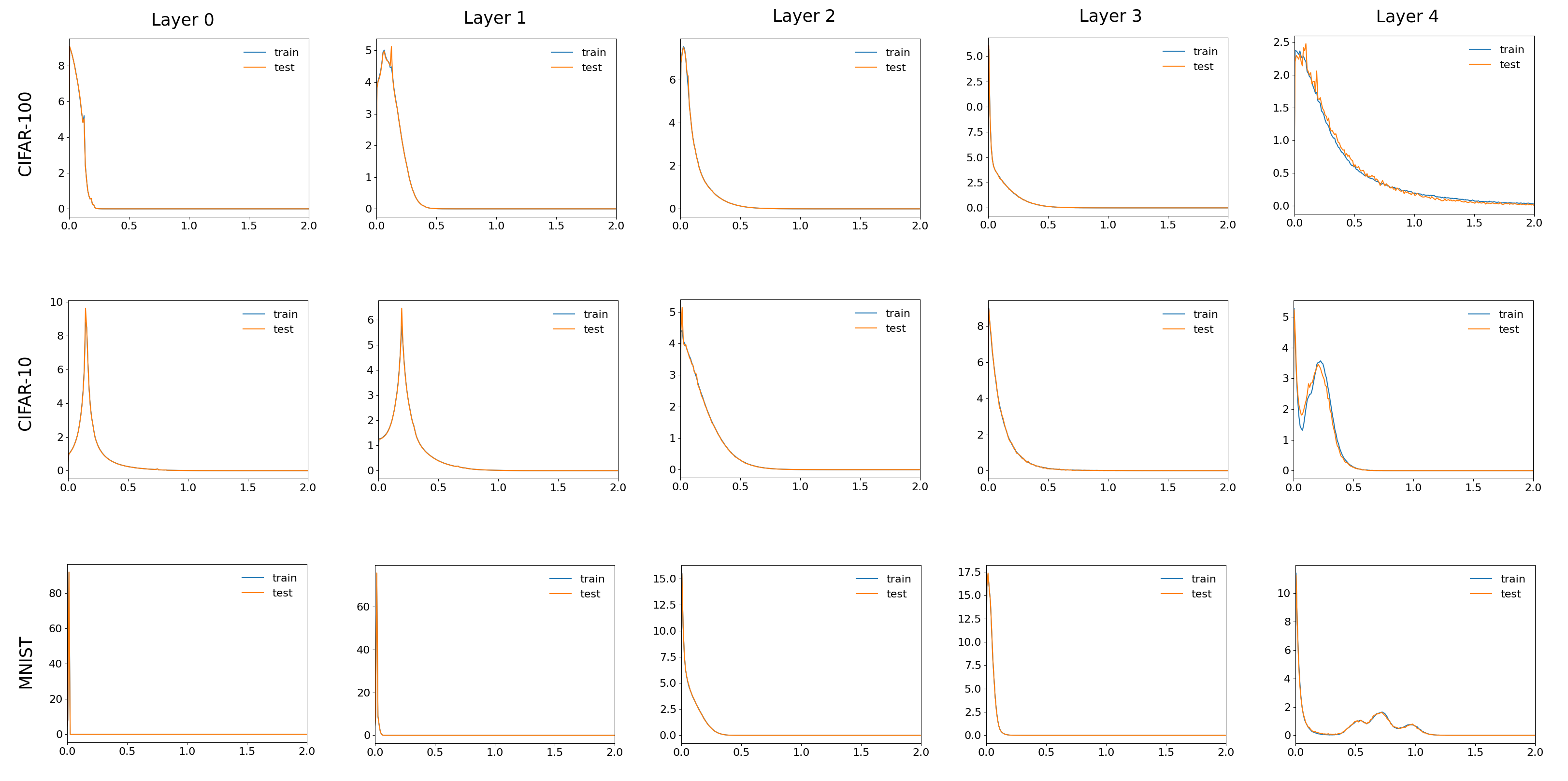}
\caption{Supplemental examples of univariate histograms for ResNet-18 for feature 1 across CIFAR-100, CIFAR-10, and MNIST}
\label{figures:marginal_extra}
\end{figure*}

These intervals are visualized as shaded bands around the mean KL-divergence values in Figures \ref{figures:fit} and \ref{figures:fit_extra}. They indicate the uncertainty in the average KL-divergence for each fitted distribution and enable statistical comparison across distributions and layers. Overlapping intervals suggest no significant difference, while non-overlapping intervals indicate a statistically significant difference in goodness-of-fit.

\section{Appendix: Experimental Design for Copula Inter-comparison}
\label{app:confidence}

This appendix provides a detailed description of the methodology used in analysis of copula interdependence.

\subsection*{Separate Processing for Training and Testing} 
    We have extracted training features from the training data and testing features from the testing data. First, we obtain the empirical marginal and interdependence terms strictly from the training data. Once we obtain these terms, we evaluate our model of copula interdependence by determining how well it fits the probability density of the withheld test features by using the criteria of cross entropy loss.

\subsection*{Probability Integral Transform}

In our implementation, the empirical PIT is calculated by sorting all of the training features in the range $[0,n-1]$ in order to obtain a set of $n$ ordered ranks.
Typically, the probability integral transform converts a marginal distribution into a uniform distribution over the interval $(0, 1)$. However, in our approach, we carry out the analysis using a rescaled version of the probability integral transform that maps the feature values to the interval $(-1, 1)$. This rescaling is motivated by the fact that many standard orthogonal functions are defined on this interval, allowing us to represent the copula density in a richer and more flexible way without parametric assumptions. The modified probability integral transform is defined in the following equation.
\begin{equation}
\begin{aligned}
F_i(x) = 2 \cdot \Pr[X_i \leq x] - 1
\end{aligned}
\end{equation}

\subsection*{Copula Density and Its Evaluation}

The empirical moments $\hat{\mu}$ and copula density $\hat{c}$ are calculated in 
a $C$ program that takes as input the entire training sample for the specified $D$ features, and outputs a set of $K^D$ empirical moments known as $\hat{\mu}$.  As such the empirical moments are computed entirely from the training set.  This set of moments further fully defines a model of the copula interdependence $\hat{c} : \mathbb{R}^D \rightarrow \mathbb{R}$ which is the dot product of the moments and the set of orthogonal functions.

\begin{equation}
\hat{c}(\vec{y}) \quad = \quad \hat{\mu} \cdot \Phi(y) \quad = \sum_{T \in \mathbb{Z}_K^D} \hat{\mu}_T \; \Phi(\vec{y})
\end{equation}

Now that our copula interdependence model is $\hat{c}$ is estimated from the training data, our task is to evaluate how well it models the probability density of the features from the test set.  Cross entropy loss is used for the evaluation criteria.

For a given set of test features $X_{test}$, the transformed test features $Y_{test}$ are calculated using the probability integral transform.  Then, we use our model $\hat{c}$ to determine the predicted probability of the test features.  This predicted probability is compared against the true probability of $1/N_{test}$ because empirically each test feature is equally likelihood.  Therefore, the overall cross entropy evaluation is calculated using the following summation over the test set.  

\begin{equation}
\begin{aligned}
\text{Cross-Entropy} \; = \; -\frac{1}{N_{\text{test}}} \; \sum_{y \in Y_{test}} \log\left(\hat{c}(y)\right)
\end{aligned}
\end{equation}

\subsection*{Confidence Intervals}

This training and testing process is repeated 30 times for each model, dataset, and layer using a different subset of $D$ features.  For this analysis we used $D=4$.  By repeating this process 30 times, we straightforwardly calculate $95\%$ confidence intervals using Student's t-test for the reported cross entropy loss statistics.

\subsection*{Archimedean copulas}

Archimedean copulas make use of a Generator function $\Psi(y; \theta)$ that is invertible as follows.

\begin{equation}
C(y_1, y_2, \theta) \quad = \quad \Psi^{-1} \big( \; \Psi(y_1;\theta) \; + \; \Psi(y_2;\theta) \; ; \; \theta \; \big)
\end{equation}

The Gumbel \citep{gumbel1960bivariate}, Frank \cite{frank1979simultaneous}, Clayton \citep{clayton1978model}, Ali Mikhail and Haq (AMH) \citep{ali1978class}, and Joe \citep{joe1994multivariate} generator functions are the most popular choices.
The hyperparameter $\theta$ was determined using a formula based on either Spearman's $\rho$ or Kendall's $\tau$ of the bivariate series.  For Gumbel, Clayton and AMH, this formula is of closed form.  For Joe and Frank the inverse formula is closed form.  For the other methods, the inverses were calculated to high precision using binary search of the following equations.

\begin{equation}
\begin{aligned}
\text{Gumbel:} & \quad \theta = \frac{1}{1-\tau} \\
\text{Clayton:} & \quad \theta = \frac{2 \tau}{1 - \tau} \\
\text{AMH:} & \quad \theta = \frac{3 \rho}{3 + \rho} \\
\text{Joe:} & \quad \tau = 1 - 4 \sum_{k-1}^\infty \frac{1}{k \big(\theta k + 2 \big) \big(\theta (k-1)+2 \big)}  \\
\text{Frank:} & \quad \tau = 1 - \frac{4}{\theta} \left( 1 - \int_0^\theta \frac{t}{e^t - 1} dt \right) \\
\end{aligned}
\end{equation}

\section{Appendix: Additional Analysis Plots}
\label{app:marginal}

\begin{figure*}[hbt]
\centering
\includegraphics[width=1.0\textwidth]{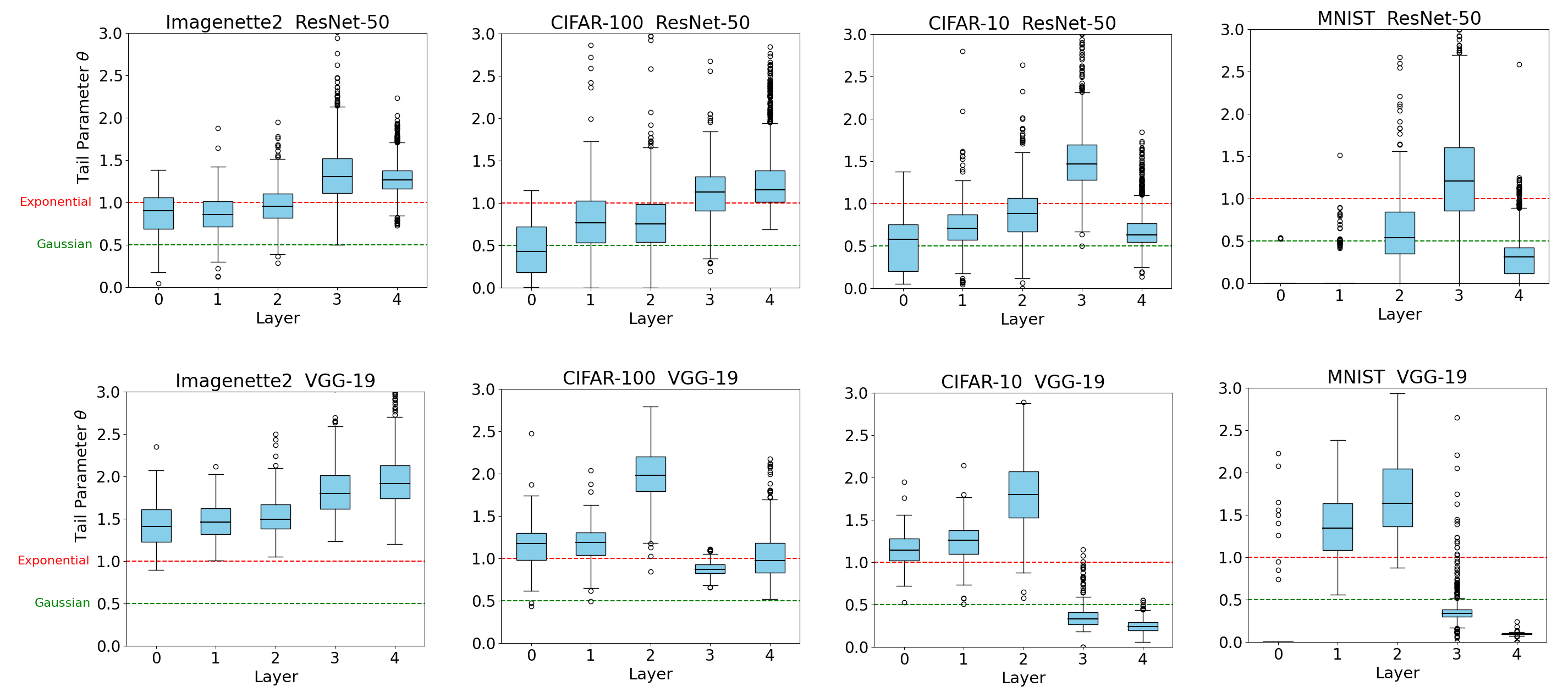}
\caption{Estimated Weibull tail parameter $\theta$ for ResNet-50 and VGG-19 features across layers. Larger values correspond to heavier-tailed activation distributions.}
\label{figures:boxplot_extra}
\vspace{10pt}
\centering
\includegraphics[width=0.7\textwidth]{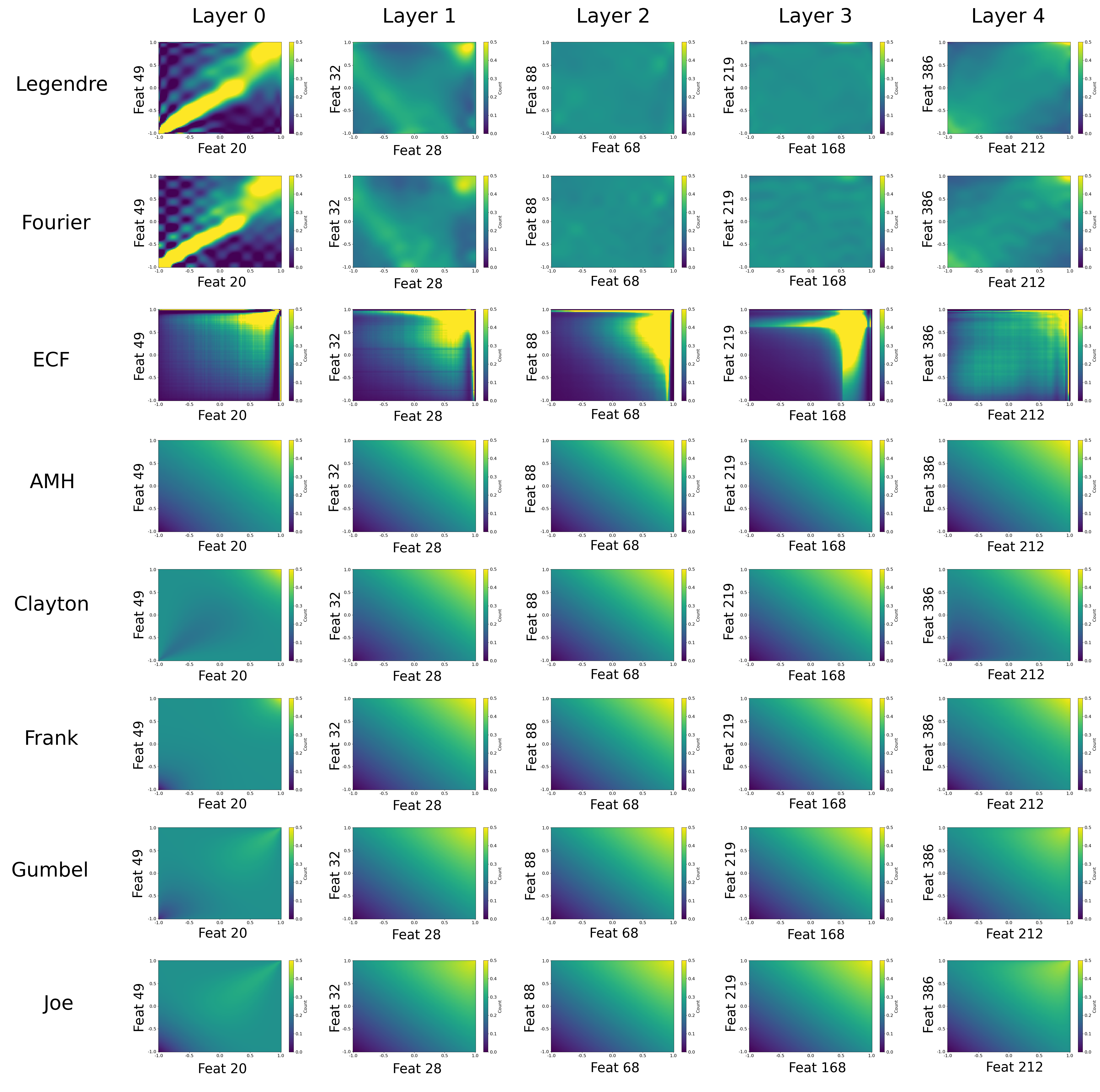}
\caption{Comparison of copula density over random bivariate features for all methods using ResNet-18 on Imagenette2.}
\label{figures:boxplot_extra}
\end{figure*}



\end{document}